# Spectral Machine Learning for Pancreatic Mass Imaging Classification

Yiming Liu[a], Ying Chen[b,c], Guangming Pan[a], Weichung Wang[d], Wei-Chih Liao[e], Yee Liang Thian[f], Cheng E. Chee[g], Constantinos P. Anastassiades[h]

[a]School of Physical & Mathematical Sciences, Nanyang Technological University, Singapore
[b]Department of Mathematics, National University of Singapore, Singapore
[c]Risk Management Institute, National University of Singapore, Singapore
[d]Institute of Applied Mathematical Sciences, National Taiwan University, Taiwan
[e]Department of Internal Medicine, National Taiwan University, Taiwan
[f]Department of Diagnostic Imaging, National University Hospital, Singapore
[g]National University Cancer Institute, Singapore, [h]Division of Gastroenterology and Hepatology, Department of Medicine, National University Hospital, , Singapore.

March 23, 2021

**Abstract**

We present a novel spectral machine learning (SML) method in screening for pancreatic mass using CT imaging. Our algorithm is trained with approximately 30,000 images from 250 patients (50 patients with normal pancreas and 200 patients with abnormal pancreas findings) based on public data sources. A test accuracy of 94.6% was achieved in the out-of-sample diagnosis classification based on a total of approximately 15,000 images from 113 patients, whereby 26 out of 32 patients with normal pancreas and all 81 patients with abnormal pancreas findings were correctly diagnosed. SML is able to automatically choose fundamental images (on average 5 or 9 images for each patient) in the diagnosis classification and achieve the above mentioned accuracy. The computational time is 75 seconds for diagnosing 113 patients in a laptop with standard CPU running environment. Factors that influenced high performance of a well-designed integration of spectral learning and machine learning included: 1) use of eigenvectors corresponding to several of the largest eigenvalues of sample covariance matrix (spike eigenvectors) to choose input attributes in classification training, taking into account only the fundamental information of the raw images with less noise; 2) removal of irrelevant pixels based on mean-level spectral test to lower the challenges of memory capacity and enhance computational efficiency, while maintaining superior classification accuracy; 3) adoption of state-of-the-art machine learning classification – gradient boosting and random forest. Our methodology showcases practical utility and improved accuracy of image diagnosis in pancreatic mass screening in the era of AI.

# 1 Introduction

Pancreatic cancer is the third leading cause of cancer death in the United States in 2021 [22]. It remains one of the few cancers with rising incidence and five-year overall survival remains less than 10%. There are no reliable screening tests for pancreatic cancer in the general population and most patients are asymptomatic until they reach an advanced stage disease at diagnosis. Pancreatic cancer progresses from pre-cancerous lesions to invasive cancer over a variable time period. The use of imaging with higher accuracy to detect pre-cancerous lesions or early cancer may improve survival.

Although computerized tomography (CT) imaging is widely used in all the stages of the clinical treatment, there is concern on the topic of accuracy and also the time needed in the imaging classification, as lesions are often subtle and the number of specialty radiologists is insufficient for burgeoning imaging demand. The sensitivity was reported to be between 76% to 92% for diagnosing pancreatic masses or abnormalities with contrast-enhanced abdominal CT scans [2],[3], while false positive diagnosis could lead to unnecessary diagnostic procedures. As a first step in image interpretation, radiologists and clinicians need to identify the relevant slices of the imaged CT volume containing pancreatic anatomy for review. This step is potentially time consuming and laborious, given the large number of thin section images that modern CT scanners output.

With the recent breakthroughs in machine learning, the use of artificial intelligence (AI) has been advocated to assist radiologists in imaging diagnosis. In lung cancer, a convolutional neural network (CNN) such as DeepScreener [19] and Deep Feature Transfer Learning [15] have been used. [18], [9],[20] and [12] For image selection, the Histogram Threshold Tail approach [17] and automatic priori knowledge method [21] have been developed for liver segmentation. Deep learning-based computer-aided diagnosis (CADx) was performed in breast and lung lesions that showed robustness to potential errors caused by inaccurate image processing [7].

Despite significant improvements, the success of the machine learning approaches depends on 1) availability of high quality training sample and 2) advanced computational facilities such as GPU and large memory capacity. It is worth mentioning that in many machine learning experiments images are directly used as input attributes, although many image pixels are likely irrelevant for diagnosis but more likely introduce noises. Compared to the high standard requirements on data quality and computing facility for successful implementations of machine learning, spectral learning is a powerful tool to extract essential information that is often low-dimensional from big data. From a signal processing perspective, spectral learning has much lower requirements. Specifically, spike eigenvectors, corresponding to the largest eigenvalues of data covariance matrix, are found to capture important structures or directions of the original big data. In spectral analysis, the spike eigenvectors are particularly effective in separating signals from noise, and simultaneously carry out dimensionality reduction as demonstrated by several groups (see examples in [16], [13] and [1]). In other words, the spike eigenvectors encode large and quality information at low memory cost.

We present a novel spectral machine learning (SML) method for the diagnosis of pancreatic masses on CT scan. 1) Specifically, for each patient, it extracts the first two spike eigenvectors of the covariance matrix of the CT images. As there are in general two body parts scanned in the public CT dataset, chest and abdomen, the two spike eigenvectors are found to well represent the original large-dimensional information of the CT scan. Compared to raw images, the spike eigenvectors are not only low dimensional, but also contain less noise. 2) In practice, each patient may have different number of images in the CT scan. A quantile comparison is thus proposed. Specifically, CT images are arranged in order according to a patient's individual spike eigenvectors. For any given constant $\alpha$, the same $\alpha$-quantile images of different patients are compared, which will be further used as input attributes to perform the state-of-the-art image classification with gradient boosting and random forest. 3) Certain types of image pixels have little contribution in distinguishing between patients with normal pancreas and patients with an abnormal pancreas mass. In other words, including these pixels in classification wastes computational memory and lowers computational efficiency. We adopt a spectral mean-difference test to identify the irrelevant image pixels among all patients. We remove the irrelevant pixels and only utilize the rest in the classification training. This further reduces the requirement for computing facilities.

Our SML algorithm is trained with approximately 30,000 images of 250 patients (50 patients with normal pancreas and 200 patients with abnormal pancreas findings) based on public data sources. It achieved a test accuracy of 94.6% in the out-of-sample diagnosis classification based on a total of approximately 15,000 images of 113 patients. Only 6 out of 32 patients with normal pancreas and none of the 82 patients with an abnormal pancreas finding were wrongly diagnosed. Moreover, the SML automatically identified a few fundamental images, on average 5 or 9 images for each patient, when achieving the high accuracy in the diagnosis classification. We acknowledge the high performance of a well-designed integration of spectral learning and machine learning, where the spike eigenvectors are utilized to determine input attributes in the gradient boosting and random forest, taking into account only the fundamental information of the raw images with less noise in classification training. It perfectly combines the advantages of the two kinds of learnings in terms of accuracy and information extraction, making an effective and stable image classification possible with low requirements on data quality and computing facilities. Our methodology showcases practical utility for future use to assist radiologists in pancreatic mass screening in the era of AI. To enhance research transparency, the code is publicly available at https://github.com/liuy0135/SML-CT.git.

## 2 Data

Our retrospective study was conducted using public data sources from *https://wiki.cancerimagingarchive.net/display/Public/Pancreas-CT* and *medicaldecathlon.com*, which contained contrast-enhanced abdominal 3D volume CT scans from 363 patients. There were two groups. The first group consisted of 82 patients from the National Institutes of Health Clinical Centre with normal pancreas on imaging, 17 of whom were healthy kidney donors scanned prior to nephrectomy and the remaining 65 patients who were scanned for other reasons and found to have neither major abdominal pathologies nor pancreatic cancer lesions. The second group had 281 patients undergoing resection of pancreatic masses (intraductal mucinous neoplasms, pancreatic neuroendocrine tumours, or pancreatic ductal adenocarcinoma). The CT scans contained 46,047 images, including 19,328 images in the group with normal pancreas (range of 181 to 481 images per patient), and 26,719 images in the group with abnormal pancreas findings (range of 37 to 751 images per patient). These CT scan images were provided by Memorial Sloan Kettering Cancer Centre (New York, NY, USA) and were previously reported in radiomic applications [4], [6]. All the CT scan slice thickness was between 1.5 to 2.5 mm for both groups. In this cohort study, the subjects' age and gender were not of any concern in image classification, and hence omitted. For more information, one can refer to the website of raw data. Moreover, we transformed all image slices into resolutions of 128×128 pixels. One may also transform the images into any other resolution, which will not affect the implementation of the proposed SML method.

To evaluate the sensitivity of pancreatic image diagnosis, we conducted cross-validation. In each iteration, we randomly selected 50 and 200 CT scans from the normal pancreas group and the group with abnormal pancreas findings, respectively, as the labelled training sample. We used the remaining CT scans, 32 from the normal pancreas group and 81 from the group with abnormal pancreas findings, as the test sample. The CT scans in the test sample were treated as no label. The true label was only utilized when evaluating the out-of-sample classification accuracy. The proposed SML method was trained to select the relevant images (and also the pixels) for each patient and perform diagnosis based on the 250 labelled patients for in-sample. Subsequently, the trained model was used to perform out-of-sample test for the remaining 113 patients. The performance was measured on average accuracy of the cross-validation experiments.

## 3 Spectral Machine Learning

CT scans contain a large amount of images with the presence of noise. We performed spectral learning to extract fundamental structure in the data as it should be robust to noisy images. In the first step, we selected essential images of all the patients to build up training sample. For each patient, we performed spectral decomposition to obtain spike eigenvectors of the covariance matrix of his/her images, where the spike eigenvectors correspond to the largest eigenvalues. The spike eigenvectors of each patient were ordered and the quantile images of CT scans were defined. The optimal quantile images were then selected for all the patients based on the training accuracy. Next, we removed the irrelevant pixels of the selected quantile images, which would need less memory space in computation. We presented a spectral mean-difference test to identify the "useless" pixels of the quantile images of all patients. Lastly, the machine learning method, i.e. gradient boosting and random forest, was used to perform classification based on the selected pixels. Details of the setup and algorithms are below.

To simplify the presentation, we introduce some notations. Suppose there are $n$ patients. We use $k$ to denote the patient ID. For each patient $k$, there are $m_k$ images of his/her CT scan. Let

$$\mathcal{X} = \{\mathcal{X}_k\}_{k=1}^n, \text{ where } \mathcal{X}_k = \{\mathbf{X}_{k,j} \in \mathbb{R}^{p \times p}\}_{j=1}^{m_k}$$

be the (training) sample of CT scans. Here, $\mathcal{X}_k$ refers to the collection of the $m_k$ images of patient $k$, where $k = 1, \cdots, n$ and $j$ represents the image ID of patient $k$. Note that $m_k$ varies for different patient $k$. Every image has the same resolution with $p \times p$ pixels. In our study, $n = 250$, $m_k$ varies from 37 to 751 for patient $k$. We transformed each image to a fixed size of 128 × 128 pixels (i.e. $p = 128$).

## 3.1 Selection of essential images via spectral learning

We first compute the spike eigenvectors for each patient. For patient $k$, one can transform each $p \times p$ image via vector operator, which stacks the columns of a matrix into a vector with $p^2$ elements. Given $m_k$ image vectors, it is easy to compute the sample covariance as follows

$$\begin{aligned} S_k = [s_{k,ij}] &= \left[\text{vec}(\mathbf{X}_{k,i})^\top \text{vec}(\mathbf{X}_{k,j})\right] \in \mathbb{R}^{m_k \times m_k} \\ &= (\mathbf{v}_1^k, \ldots, \mathbf{v}_{m_k}^k)^\top \begin{pmatrix} \lambda_1^k & \cdots & 0 \\ \vdots & \ddots & \vdots \\ 0 & \cdots & \lambda_{m_k}^k \end{pmatrix} (\mathbf{v}_1^k, \ldots, \mathbf{v}_{m_k}^k) \end{aligned} \quad (1)$$

which gives an $m_k \times m_k$ covariance matrix. By decomposition, one obtains $m_k$ eigenvectors $\mathbf{v}_j^k$ corresponding to $m_k$ descending eigenvalues $\lambda_j^k$ with $j = 1, \cdots, m_k$. The eigenvectors point the direction of variations in the images. Each element in the spike eigenvector is associated with one original image projected in the direction. Therefore, sorting an eigenvector in e.g., increasing order corresponds to ordering images, as the linear mapping in decomposition does not change its sequence.

The initial step in imaging interpretation for radiologists and computer aided diagnosis algorithms is to localise the relevant images in $\mathcal{X}_k$ before abnormality detection. This requires experience and expert domain knowledge. We now propose a spectral learning approach that is to perform a purely data driven image selection.

We use $j_k$ to denote the image ID for patient $k$ and the aim is to choose a set of images $\mathcal{Y}_k = \{\mathbf{Y}_{k,1} \ldots, \mathbf{Y}_{k,j_k} \in \mathbb{R}^{p \times p}\} \subset \mathcal{X}_k$ containing the essential images to be used for classification. The spike eigenvectors of the covariance matrix $S_k$ are known to demonstrate certain fundamental structure of data, i.e. images $\mathcal{X}_k$ in our case. If there are two clusters with different means, [14] shows that the first two spike eigenvectors corresponding to the first two largest eigenvalues, after sorting in an increasing order, at least one of them displays a gap between the two distinct clusters. In our case, it observes that there is always a jump in one of the two ordered spike eigenvectors, implying there exists at least two clusters in one CT scan. This is consistent to the fact that the images of chest and abdomen parts are different. Thus it motivates consideration of the first two spike eigenvectors only in the spectral learning. Numerically this leads to two sets of sorted images for any patient $k$, associated with the first and second spike eigenvectors. Mathematically, we denote by $\mathbf{v}^k_{(\ell)}$ the $\ell$-th sorted eigenvector, where $\mathbf{v}^k_\ell$ is the unsorted eigenvector corresponding to $\ell$-th largest eigenvalue of $S_k$. Note that we have $\ell = 1, 2$, where the two clusters refer to the two body parts of a patient. It shouldn't be mixed with the two groups of patients with distinct pancreatic conditions.

If the number of CT images of each patient is same, one can naturally use the ordered images for a comparison, which is similar to compare part by part for all patients. However, the number of images differs and there are individual $m_k$ images for each patient, making the one-to-one direct comparisons impossible. A study by [14] shows that the magnitudes of the elements in the spike eigenvectors are associated with the mean of the images. Given that the eigenvector has a unit norm, it means that the quantiles of spike eigenvectors are appropriate for one-to-one comparison for different patients. In other words, one can perform classification based on the selected quantile images.

Given a sorted spike eigenvector $\mathbf{v}^k_{(\ell)}$ (rank $\mathbf{v}^k_\ell$ from the smallest to the largest), the $\alpha$-th quantile, with $\alpha \in [0,1]$, of the spike eigenvector is corresponding to one specific image of patient $k$, which is named quantile image. Since both $\mathbf{v}^k_\ell$ and $-\mathbf{v}^k_\ell$ are the $\ell$-th eigenvector of $S_k$, the sign of $\mathbf{v}^k_\ell$ may affect the ranking of $\mathbf{v}^k_\ell$. To eliminate the sign effect, we propose a standard normalization approach to order the eigenvector, see Algorithm 2. Given any $\alpha$, we choose the $\alpha$-th quantile image for all patients $k = 1,...,n$. We define an $\alpha$-th indicator for the quantile image:

$$i^{(\ell)}_{\alpha,k} = \arg \min_{i \in \{1,...,m_k\}} \left| \mathbf{v}^k_{\ell,i} - \mathbf{v}^k_{(\ell)}(\alpha) \right|, \tag{2}$$

and define $\mathbf{X}_{k, i^{(\ell)}_{\alpha,k}} \in \mathcal{X}_k$ to be the $\alpha$-th quantile image. For a fixed $\alpha$ and $\ell$, we obtain, among the $n$ patients, a set of quantile images

$$\mathcal{T}_{\alpha,\ell} := \left\{ \mathbf{X}_{1, i^{(\ell)}_{\alpha,1}}, ..., \mathbf{X}_{n, i^{(\ell)}_{\alpha,n}} \in \mathbb{R}^{p \times p} \right\}.$$

The quantile level $\alpha$ is a constant determined by user. We also provide a data-driven way to choose the optimal value of $\alpha$ for good image classification. Specifically, we consider a set of regularly spaced grids of $\alpha = 0, 0.02,...,0.98, 1$, where $\alpha = 0$ means one considers the image corresponding to the smallest element in the spike vector, and $\alpha = 1$ refers to the images associated with the largest element in the spike vector. In a pre-processing step, we perform simple classification among $n$ quantile images, say e.g. K-means, for each $\alpha$. Then choose some values of $\alpha$ with the smallest misclassification errors. The multiple data-driven selected quantile images form the training sample of $\mathcal{T}_{\alpha,\ell}$.

## 3.2 Removal of irrelevant pixels

For any pre-specified quantile $\alpha$, there are $n$ images of all patients $\mathcal{T}_{\alpha,\ell} := \{\mathbf{X}_{1,i_{\alpha,1}}, \dots, \mathbf{X}_{n,i_{\alpha,n}} \in \mathbb{R}^{p \times p}\}$ belonging to two groups (pancreas normal and abnormal groups). We label the two groups by $\mathcal{C}_1$ and $\mathcal{C}_2$. Before proceeding with classification training, it is necessary to distinguish image pixels with different contribution in classification and remove the useless pixels to save computational cost. For notational simplification, we omit the subscript $i_{\alpha,k}$ of $\mathbf{X}_{k,i_{\alpha,k}} \in \mathbb{R}^{p \times p}$. Write $\mathbf{Z}_k = \text{vec}(\mathbf{X}_k) \in \mathbb{R}^{p^2}$ with $k = 1, \cdots, n$ as the vectorized $\mathbf{X}_k$, and denote the $i$-th coordinate of $\mathbf{Z}_k$ by $\mathbf{Z}_{k,i}$. Here, $\mathbf{Z}_{k,i}$ corresponds to one pixel, i.e. the $i$-th pixel of the $k$-th image.

Specifically, we assign each pixel of image into one of three exclusive sets. The first set, denoted as $\mathcal{A}_1$, contains pixels where all $\mathbf{Z}_{k,i}$ with $k = 1, \cdots, n$ share the same pixel value, namely $\mathbf{Z}_{k_1,i} = \mathbf{Z}_{k_2,i} = Constant$ for any $k_1 \neq k_2$. As an example, we discuss the grouping of $\mathbf{Z}_{k,1}$, i.e. the first coordinate of $\mathbf{Z}_k$ with $i = 1$ to illustrate the idea. If the pixel belongs to $\mathcal{A}_1$, obviously the first coordinate is useless as all patients have the same values of $Z_{k,1}$. Thus, we may safely remove this pixel from the following analysis. The second set, denoted as $\mathcal{A}_2$, has different pixel values of $\mathbf{Z}_{k,i}$ among the $n$ patients. However, the difference is purely caused by randomness. Consider the first coordinate again. Let $\mu(\mathbf{Z}_{k,1})$ and $\sigma(\mathbf{Z}_{k,1})$ be the mean and variance of the first coordinate, respectively. This case $\mu(\mathbf{Z}_{k_1,1}) = \mu(\mathbf{Z}_{k_2,1})$ and $\sigma(\mathbf{Z}_{k,1}) \neq 0$ with all $1 \leq k, k_1, k_2 \leq n$. In other words, set $\mathcal{A}_2$ implies that all observations $\left(\{\mathbf{Z}_{k,1}\}_1^n\right)$ have the same mean, even for different groups. The different pixel values are random merely drive by the nonzero variance. We will show soon that this category is also irrelevant for classification and should be removed too. Set $\mathcal{A}_3$ is the complementary set of the aforementioned two cases.

The grouping is implemented for each pixel $\mathbf{Z}_{k,i}$ for $i = 1, \cdots, p^2$. Formally, we divide the $1, \cdots, p^2$ coordinates of $\mathbf{X}_{k,i}$ (or pixels of image) into 3 groups as explained above: coordinates with the same value among all the $n$ images, denoted by $\mathcal{A}_1$; coordinates with the same mean but different pixel values, denoted by $\mathcal{A}_2$; and the rest coordinates, denote by $\mathcal{A}_3$

$$\begin{aligned}
\mathcal{A}_1 &= \{i: \mathbf{Z}_{k_1,i} = \mathbf{Z}_{k_1,i} \text{ for all } 1 \leq k_1, k_2 \leq n\} \\
\mathcal{A}_2 &= \{i: \mu(\mathbf{Z}_{k_1,i}) = \mu(\mathbf{Z}_{k_2,i}) \text{ for all } 1 \leq k_1 \neq k_2 \leq n\} \setminus \mathcal{A}_1, \\
\mathcal{A}_3 &= \{1, \dots, p^2\} \setminus (\mathcal{A}_1 \cup \mathcal{A}_2)
\end{aligned} \tag{3}$$

To specify $\mathcal{A}_1$, we estimate

$$\hat{\mathcal{A}}_1 := \left\{ 1 \leq i \leq p^2 : \frac{1}{n} \sum_{k=1}^n \left( \mathbf{Z}_{k,i} - \bar{\mathbf{Z}}_{\cdot,i} \right)^2 < (2\log(n))^{-1} \right\},$$

where $\mathbf{Z}_{\cdot,i} = \sum_{k=1}^{n} \mathbf{Z}_{k,i}/n$ and $(2\log(n))^{-1}$ measures statistical significance. To identify $\mathcal{A}_2$, we compute the normalized variance difference between the two groups $\mathcal{C}_1$ and $\mathcal{C}_2$:

$$\mathbf{d}_i := \left(\frac{1}{n_1}\sum_{k\in\mathcal{C}_1} \mathbf{Z}_{k,i} - \frac{1}{n_2}\sum_{k\in\mathcal{C}_2} \mathbf{Z}_{k,i}\right)^2 / \mathbf{s}$$

where $\mathbf{s} = [(n_1-1)\mathbf{s}_1 + (n_2-1)\mathbf{s}_2]/(n-2)$, $\mathbf{s}_j$ is the sample variance under $\mathcal{C}_j$ with $j = 1,2$, $n_1$ and $n_2$ are the cardinality of $\mathcal{C}_1$ and $\mathcal{C}_2$, respectively. Note that the class labels $\mathcal{C}_1$, $\mathcal{C}_2$ are known for training sample. Hence, $\mathbf{d}_i$ is computable and reflects the difference of the $i$-th coordinate between $\mathcal{C}_1$ and $\mathcal{C}_2$. Set

$$\hat{\mathcal{A}}_2 := \{i \in \{1, \ldots, p^2\} \setminus \hat{\mathcal{A}}_1 : \mathbf{d}_i < (\log(n))^{-1}\} \text{ and } \hat{\mathcal{A}}_3 = \{1, \ldots, p^2\} \setminus (\hat{\mathcal{A}}_1 \cup \hat{\mathcal{A}}_2)$$

where $(\log(n))^{-1}$ measures statistical significance. Under mild conditions, the sets $\hat{\mathcal{A}}_1$ and $\hat{\mathcal{A}}_2$ are the consistent estimators of $\mathcal{A}_1$ and $\mathcal{A}_2$, respectively.

Table 1 reports the percentage of image pixels belonging to the three sets $\mathcal{A}_1, \mathcal{A}_2$ and $\mathcal{A}_3$ for some $\alpha$'s when $\ell = 2$. In the training sample, 21.3% pixels belong to $\mathcal{A}_1$, which have the same values and thus useless and can be safely removed without any loss. The set $\mathcal{A}_2$ contains 17.1% ~ 41.0% pixels, whose mean difference is statistically insignificant among the $n$ images, and thus irrelevant for classification too. After the removal, only the pixels of set $\mathcal{A}_3$, 37.7% ~ 61.6% of image pixels, will be used for classification training. In other words, almost half of the raw pixels can be removed. Moreover, the size of $\mathcal{A}_3$ generally decreases with $\alpha$, indicating that smaller $\alpha$'s contain more information for classification.

Table 1: Percentage (%) of image pixels in $\mathcal{A}_1, \mathcal{A}_2$ and $\mathcal{A}_3$ for some $\alpha$'s.

|  | $\alpha = 0$ | $\alpha = 0.2$ | $\alpha = 0.4$ | $\alpha = 0.6$ | $\alpha = 0.8$ | $\alpha = 1$ | Avg. |
|---|---|---|---|---|---|---|---|
| $\mathcal{A}_1$ | 21.3 | 21.3 | 21.3 | 21.3 | 21.3 | 21.3 | 21.3 |
| $\mathcal{A}_2$ | 17.1 | 25.6 | 37.4 | 39.6 | 38.4 | 41.0 | 33.2 |
| $\mathcal{A}_3$ | 61.6 | 53.1 | 41.3 | 39.1 | 40.3 | 37.7 | 45.5 |

### 3.3 Gradient boosting and random forecast

Known as efficient machine learning approaches, Gradient Boosting (GB) and Random Forest (RF) have delivered promising accuracy results in various kinds of classification analysis. RF is an ensemble learning method, which is constructed by a multitude of decision trees [10],[11],[8]. GB builds model in a stage-wise fashion as other boosting methods do [5], which is typically used with decision trees of a fixed size as base learners [8]. In our study, we combined GB with RF and performed image classification.

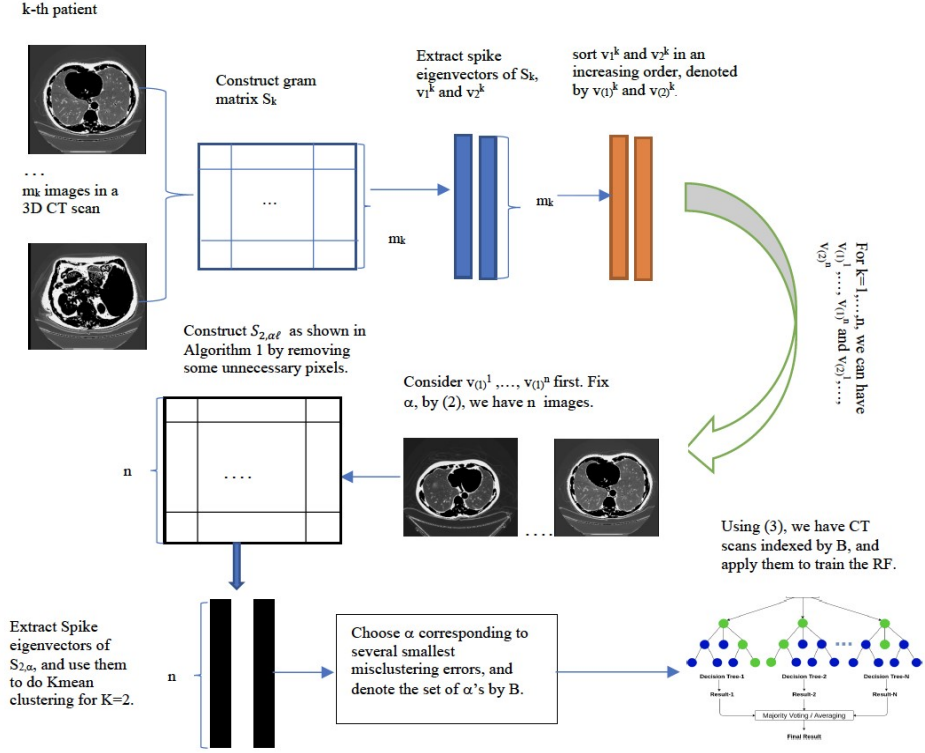

Figure 1: The overview of the proposed method.

The input attributes of the machine learning methods are the relevant pixels of the selected quantile images, i.e. the pixels in $\mathcal{A}_3$ for the selected $\alpha$'s. When training random forest, we use the mean of the selected quantile images, for a robust prediction. According to the concentration property, the sample mean is a consistent estimator of the true mean under mild conditions. Compared to each individual image, the mean image smooths out errors and removes noises. The random forest is constructed with 1,000 trees and 10 or 20 randomly chosen features for each tree.

Figure 1 demonstrates the flowchart of the proposed spectral machine learning (SML) imaging classification method. It is an integration of spectral learning and machine learning. The spectral learning is used to select relevant attributes given large and noisy data, while the machine learning is adopted to perform classification. The algorithms can be formulated as follows:

**Algorithm 1**: Algorithm for SML

**Training part**

Given n CT scans from n patients, i.e., $\mathcal{X} = \{\mathcal{X}_k\}_{k=1}^n$, where $\mathcal{X}_k = \{\mathbf{X}_{k,j} \in \mathbb{R}^{p \times p}\}_{j=1}^{m_k}$ as a testing set.

For each $\mathcal{X}_k = \{\mathbf{X}_{k,j} \in \mathbb{R}^{p \times p}\}_{j=1}^{m_k}$, where k=1,...,n do

1. Construct $S_k = [s_{k,ij}] = [\text{vec}(\mathbf{X}_{k,i})^\top \text{vec}(\mathbf{X}_{k,j})] \in \mathbb{R}^{m_k \times m_k}$, and extract the eigenvectors corresponding to largest two eigenvalues of $S_k$, denoted by $\mathbf{v}_1^k$ and $\mathbf{v}_2^k$.
2. Obtain the sorted $\mathbf{v}_{(1)}^k$ and $\mathbf{v}_{(2)}^k$ by applying Algorithm 2.

Hence, $\mathbf{v}_{(1)}^k$ and $\mathbf{v}_{(2)}^k$ for $k = 1, \ldots, n$ can be obtained.

For a given $\alpha \in [0,1]$ and a fixed $\ell$ do

3. Use (2) to obtain $i_{\alpha,k}^{(\ell)}$, and hence $\mathcal{T}_{\alpha,\ell}$ can be obtained.
4. Remove unnecessary pixels in $\mathcal{T}_{\alpha,\ell}$.
5. Construct $S_{2,\alpha\ell}$ as $S_{2,\alpha\ell} = \left[\mathbf{Z}_{i,\hat{\mathcal{A}}_2}^\top \mathbf{Z}_{j,\hat{\mathcal{A}}_2}\right]_{1 \le i,j \le n}$, and use the first two eigenvectors of $S_{2,\alpha\ell}$ to do K-means clustering.
6. Choose several $\alpha$'s and $\ell$ corresponding to the smallest misclustering error.
7. Applying the selected $\alpha$'s and $\ell$ to (2), one can have $\mathcal{Y}_k = \{\mathbf{Y}_{k,1}, \ldots, \mathbf{Y}_{k,j_k}\}$ for $k = 1, \ldots, n$. Here we set $j_k = 5$ or 9.
8. Let $\bar{\mathbf{Y}}_k = \sum_{l=1}^{j_k} \mathbf{Y}_{k,l}/j_k$, and use $\{\bar{\mathbf{Y}}_1, \ldots, \bar{\mathbf{Y}}_n\}$ to train the GB based random forest.

**Testing part**

Given m CT scans from m patients, i.e., $\widetilde{\mathcal{X}} = \{\widetilde{\mathcal{X}}_k\}_{k=1}^m$.

9. Follow same procedure as in Step 1 and 2 to obtain the sorted eigenvectors.
10. Use the selected $\alpha$'s and $\ell$ to obtain $\{\bar{\widetilde{\mathbf{Y}}}_1, \ldots, \bar{\widetilde{\mathbf{Y}}}_m\}$ as in Step 7 and 8 for the testing set.
11. Use $\{\bar{\widetilde{\mathbf{Y}}}_1, \ldots, \bar{\widetilde{\mathbf{Y}}}_m\}$ to do test by the GB based random forest.

---

**Algorithm 2**: Sort eigenvectors

1. Given a eigenvector $\mathbf{v} = (v_1, \ldots, v_m) \in \mathbb{R}^m$, treat $(v_1, \ldots, v_m)$ as the values in y-axis and $(1, \ldots, m)$ as the values in x-axis (we thus can have a plot in a plane)
2. Rescale the range of $x$ and $y$ -axis into 0 to 1 ,i.e., using $v_j - min\{v_i\})/(max\{v_i\} - min\{v_i\}$ to replace original $v_j$.
3. Set the sign of **v** to be +1 when its area under the curve (AUC) larger than 0.5, otherwise to be -1.

Given the trained classification model, we continued performing out-of-sample classification for the test sample. Specifically, for each patient in the test sample, we chose the relevant quantile images and fed in the trained random forest.

## 4 Results

We learned the parameters of SML with the training sample of 50 patients with normal pancreas and 200 patients with abnormal pancreas findings on CT. The trained model was then used to perform out-of-sample classification based on the rest, i.e. test sample with 32 patients with normal pancreas and 81 patients with abnormal pancreas findings on CT. To avoid the impact of random selection in the training/test samples, we performed cross-validation, namely iteratively chose 250 training samples with the same composition and test the rest. The performance is measured using the average of all the experiments for a robust comparison.

We also considered three alternative classification methods and implemented on the same data sets: CNN (convolutional neural network), $GRF_R/CNN_R$ (gradient boosting and random forest/CNN based on randomly selected images) and $GRF_M/CNN_M$ (gradient boosting and random forest/CNN based on the mean images of patients). The mean image was obtained by taking the average of all CT images for each patient. In the random forest, we specified 1000 trees and 10 features to be randomly selected for each tree. We evaluated the SML method and the alternatives in terms of classification accuracy for out-of-sample and computational time. All the calculations were completed under a standard CPU laptop, with IOS system, 2.3 GHz Quad-Core Intel Core i5 and memory of 8 GB 2133 MHz LPDDR3.

The performance depended on the choice of quantile $\alpha$. We used K-means clustering as it was computationally cheap and fast, though usually a suboptimal way compared to other machine learning methods such as random forest and CNN. We chose $\alpha^* = \alpha = 0$ whose misclustering error was the smallest value $0.198$. The quantile images were further used in classification. Note that the selected images were neither the first nor the last image in the CT scan, but arranged by order corresponding to the smallest element in the spike eigenvectors. Nevertheless, given that K-means has relative weak performance and the optimal quantile image only may lead to some information loss, we suggest choosing a few more quantile images, instead of only one, with small misclustering errors as input attributes. In our case, we suggest 5 or 9 quantile images up to 2% based on the second spike eigenvector, with $\alpha = 0, 0.005,...,0.02$ (grid step $0.005$) or $\alpha = 0, 0.0025,...,0.02$ (grid step $0.0025$). Details of the pre-processing are displayed in Appendix.

With the help of the spectral learning, we followed the procedure to construct the training and test samples. Table 2 reports the in-sample and out-of-sample classification accuracy when only one image was used, based on quantile image selected with SML given a fixed value of $\alpha$ (note $\alpha^* = \alpha = 0$ was the optimal choice with data-driven approach in SML), a randomly selected image ($GRF_R/CNN_R$) and the mean image ($GRF_M/CNN_M$). The SML and CNN were implemented. In SML, 10 features were used in the gradient boosting and random forest. All the results were computed based on 50 cross-validation cases. It showed that SML was superior with perfect classification for in-sample, while the best performing CNN reached 83.7%. The superior performance continued in out-of-sample test accuracy. SML reached 93.6%, a 14% absolute improvement against the best performing CNN (79.6%), 7.8% above $GRF_R$ and 10.4% above $GRF_M$.

Table 2: In-sample and out-of-sample test accuracy (%) with different $\alpha$.

|  | $\alpha$ | 0 ($\alpha^*$) | 0.2 | 0.4 | 0.6 | 0.8 | 1 | random image (R) | mean image (M) |
|---|---|---|---|---|---|---|---|---|---|
| SML | In | **100** | 100 | 100 | 100 | 100 | 100 | 100 | 100 |
|  | Out | **93.6** | 86.7 | 83.2 | 85.8 | 87.6 | 81.4 | 85.8 | 83.2 |
| CNN | In | **83.7** | 77.4 | 73.7 | 73.7 | 75.6 | 74.3 | 71.4 | 76.4 |
|  | Out | 78.8 | **79.6** | **79.6** | 71.7 | 72.6 | 70.8 | 71.7 | 71.7 |

Moreover, SML and CNN were also implemented with different quantile images to test the robustness in terms of choice $\alpha$. It showed that SML provided a robust improvement in accuracy compared to CNN for both in-sample and out-of-sample. Figure 2 displays the ROC (Receiver Operating Characteristics) curves of the SML and CNN approaches, respectively. It shows that not only the SML with the optimal quantile image (i.e. $\alpha^* = 0$) has the largest AUC (area under curve), it also has better AUC than CNN for different values of $\alpha$.

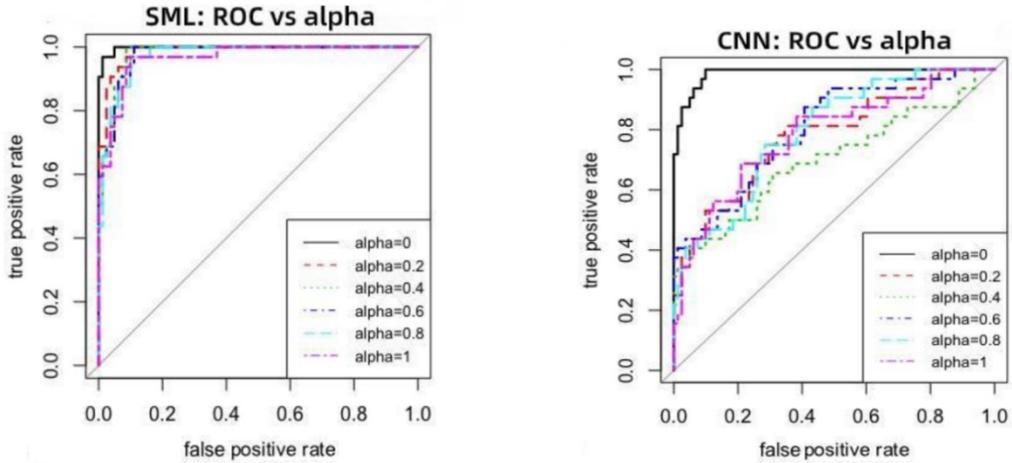

Figure 2: ROC curves by SML and CNN with different $\alpha$'s.

To further investigate the robust performance of SML with respect to the choice of quantile images, as well as the setup of the gradient boosting and random forest, we performed SML based on the average of 5 or 9 quantile images at $\alpha = 0, 0.005,...,0.02$ or $\alpha = 0, 0.0025,...,0.02$ and either 10 or 20 features in random forest. Table 3 reports the out-of-sample test classification accuracy of SML with a different setup. The in-sample accuracy was always 100% and thus omitted. Again all calculations were performed based on 50 cross-validations. It showed that more features and more image information may improve the classification accuracy, though marginal. When the mean image of the 9 quantile images and 20 features were used, the SML further improved accuracy to 94.6%.

Meanwhile, the computation cost was very low. In our case, the diagnosis operation time required 75 to 108 seconds for 113 patients. Given that the computing facility is far from fancy, it can be easily implemented on mobile applications or wearable devices. Moreover, computing time can be further shortened with a GPU server.

Table 3: Operation time (seconds) and out-of-sample test accuracy with different setup.

|  | Avg. of 5 quantile images | | Avg. of 9 quantile images | |
|---|---|---|---|---|
| Features | 10 | 20 | 10 | 20 |
| **Test accuracy** | 90.6 | 93.1 | 91.2 | **94.6** |
| Training time (sec) | 678 | 766 | 659 | 711 |
| **Test time (sec)** | 75 | 108 | 76 | 106 |

# 5  Discussion and conclusion

Our study results demonstrated the ability of spectral machine learning approach to identify pancreas abnormality, in an accurate and efficient way. The spectral machine learning classification performance on the data set was better than that of the state-of-the-art CNN approach. This finding is important because the proposed approach needs only moderate computing facility, whereas the existing deep learning requires high performance computer facility. The higher accuracy of the spectral machine learning approach was achieved by combining the spectral analysis and the gradient boosting based random forest. The proposed model is potentially applicable to other images of diseases for which training data are scarce or high performance computers are not available. To our knowledge, the model is the first attempt to classify pancreas abnormality in diagnostic imaging through the integration of spectral and machine learning analysis.

However, this study has some limitations. First, the data set was relatively small compared with conventional data sets in deep learning; in particular, the number of patients and their CT scans was moderate or even small. Second, we used a public dataset which was released for developing algorithms rather a designed study with specific clinical objectives. There is limited information on the experimental details including e.g. the equipment specification. To evaluate the robustness and generalizability of the proposed model, further investigation should be conducted in future research with larger data sets from various institutions and clearly specified clinical objectives. Third, the study was conducted for pancreas abnormality classification only. The proposed method nevertheless has potential to be used for other diagnostic imaging tasks such as detecting the location of a lesion. This needs to be pursued in future.

In conclusion, we developed a spectral machine learning model for analysis of pancreas abnormality on diagnostic imaging. There are three main contributions from this paper. First, to handle numerous CT scans, we introduced spectral analysis to select the comparable images among each CT scan, at quantile level, which is helpful in reducing computational complexity. The selection improved the diagnostic accuracy compared to other methods such as using an arbitrary or average image where the latter used all the images. Second, to relax the requirement of large memory space, which is necessary in conventional image analysis, we proposed the step of removal of irrelevant pixels, based on statistical testing, which dramatically eased computational complexity and simultaneously avoided generating the useless decision trees in the iteration of random forest, improving computational efficiency. When implemented in public data sources, the SML achieved a test accuracy of 94.6% in the out-of-sample diagnosis classification based on a total of approximately 15,000 images of 113 patients. Our methodology showcases practical utility and improved accuracy of image diagnosis in pancreatic mass screening in the era of AI.

# Appendix

**Data-driven choice of quantile images**

We chose $\alpha$ and further the quantile images based on the in-sample accuracy. We used K-means clustering as it was computationally cheap and fast, though usually a suboptimal way compared to other machine learning methods such as random forest and CNN. Figure 3 displays the plots of the misclustering error rates for 51 different values of $\alpha = [0, 0.02, \cdots, 0.98, 1.00]$ with a constant grid step of 0.02. For the first and second spike eigenvectors of one arbitrarily selected training sample, the misclustering error rates were in general smaller for the second spike eigenvector with $\ell = 2$. It also showed that the misclustering error rates monotonically increased with $\alpha$. In other words, $\alpha^* = \alpha = 0$ with the smallest misclustering error of 0.198 was the best choice. A similar pattern was observed in other training samples and thus omitted here.

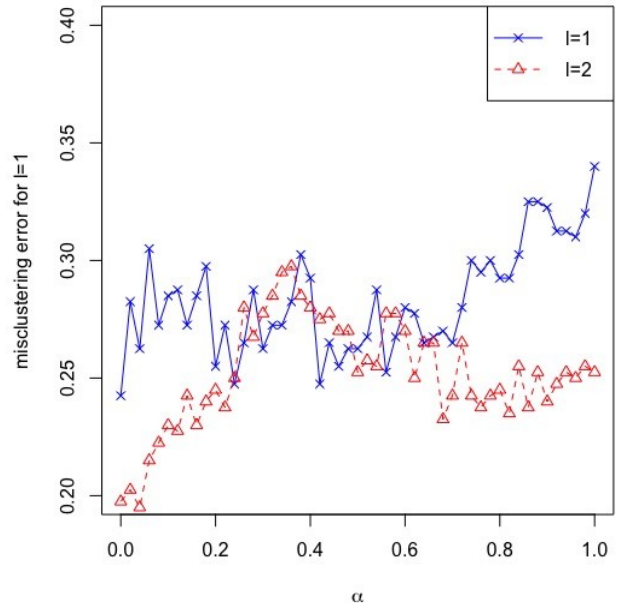

Figure 3: Fitted misclustering errors for distinct $\alpha$'s, left panel is $\ell = 1$, right panel is $\ell = 2$.